\documentclass[11pt,a4paper]{article}
\usepackage[hyperref]{emnlp-ijcnlp-2019}
\usepackage{times}
\usepackage{latexsym}
\usepackage{latexsym}
\usepackage{times}
\usepackage[export]{adjustbox}
\usepackage{amsmath, amssymb}
\usepackage[whole]{bxcjkjatype}
\usepackage{tabularx}
\usepackage{multirow}
\usepackage{bm}
\usepackage{latexsym}
\usepackage{xcolor}
\usepackage{subfigure}
\usepackage{float}
\usepackage{url}
\usepackage{courier}

\usepackage{url}

\aclfinalcopy 

\newcommand{\bhline}[1]{\noalign{\hrule height #1}}

\title{A Large-Scale Multi-Length Headline Corpus for Analyzing Length-Constrained Headline Generation Model Evaluation}

\author{Yuta Hitomi\textsuperscript{~1}~~~~~~~Yuya Taguchi\textsuperscript{~1}~~~~~~~Hideaki Tamori\textsuperscript{~1}~~~~~~~Ko Kikuta\thanks{This work was done at Retrieva, Inc. within Project.}\textsuperscript{~~2}~~~~~~~Jiro Nishitoba\textsuperscript{~2}\\ 
\textbf{Naoaki Okazaki\textsuperscript{~3}~~~~~~~Kentaro Inui\textsuperscript{~4}~~~~~~~Manabu Okumura\textsuperscript{~3}}\\
\textsuperscript{1} The Asahi Shimbun Company, \textsuperscript{2} Retrieva, Inc., \textsuperscript{3} Tokyo Institute of Technology\\
\textsuperscript{4} Tohoku University, RIKEN Center Advanced Intelligence Project\\
{\tt \{hitomi-y1, taguchi-y2, tamori-h\}@asahi.com, kikutakou@gmail.com}\\
{\tt jiro.nishitoba@retrieva.jp, okazaki@c.titech.ac.jp} \\
{\tt inui@ecei.tohoku.ac.jp, oku@pi.titech.ac.jp}
}

\date{}

\begin{document}
\maketitle
\begin{abstract}
Browsing news articles on multiple devices is now possible. The lengths of news article headlines have precise upper bounds, dictated by the size of the display of the relevant device or interface. Therefore, controlling the length of headlines is essential when applying the task of headline generation to news production. 
However, because there is no corpus of headlines of multiple lengths for a given article, previous research on controlling output length in headline generation has not discussed whether the system outputs could be adequately evaluated without multiple references of different lengths.
In this paper, we introduce two corpora, which are {\bf J}apanese {\bf N}ews {\bf C}orpus (JNC) and {\bf JA}panese {\bf MU}lti-{\bf L}ength Headline Corpus (JAMUL), to confirm the validity of previous evaluation settings.
The JNC provides common supervision data for headline generation.
The JAMUL is a large-scale evaluation dataset for headlines of three different lengths composed by professional editors.
We report new findings on these corpora; for example, although the longest length reference summary can appropriately evaluate the existing methods controlling output length, this evaluation setting has several problems. 

\end{abstract}

\section{Introduction}

\begin{table}[t]
\centering
\small
\begin{tabular}{p{7.2cm}c}
\bhline{1pt}
{\bf Article: }トヨタ自動車は18日、エンジン車だけの車種を2025年ごろまでにゼロにすると発表した。\dots ハイブリッド車~(HV)やプラグインハイブリッド車~(PHV)、燃料電池車~(FCV)も加えた「電動車」を、すべての車種に設定する。\dots \\
On the 18th, Toyota announced that it would set the model of only engine cars to zero by about 2025.\dots They set ``electric vehicle'' which is Hybrid Vehicle (HV), Plug-in Hybrid Vehicle (PHV), and Fuel Cell Vehicle (FCV) to all models.\dots\\
\bhline{1pt}
{\bf Headline for print media: }
トヨタ、全車種に電動車　25年ごろまでに　HVやFCV含め \\
All Toyota models will contain electric vehicles including HV and FCV by about 2025. \\
\bhline{1pt}
{\bf Multi-length headlines for digital media: } \\
\begin{tabular}{cp{5.0cm}}
9 chars & \textcolor{red!70!magenta}{全}車種に\textcolor{red!70!magenta}{「電動車」}\\
(10char-ref) & \textcolor{red!70!magenta}{``Electric cars''} for \textcolor{red!70!magenta}{all} models\\
13 chars & トヨタ、\textcolor{red!70!magenta}{全}車種に\textcolor{red!70!magenta}{「電動車」} \\
(13char-ref) & \textcolor{red!70!magenta}{``Electric cars''} for \textcolor{red!70!magenta}{all} Toyota's models\\
\multirow{4}{*}{ \shortstack{24 chars\\(26char-ref)} } & トヨタ、\textcolor{green!40!blue}{エンジン車だけの}車種\textcolor{green!40!blue}{ゼロへ　2025年ごろ}\\
& Toyota \textcolor{green!40!blue}{sets the number of} models \textcolor{green!40!blue}{with only engine} cars \textcolor{green!40!blue}{to zero by about 2025}.  \\
\end{tabular} 
\\
\bhline{1pt}
\end{tabular}
\caption{An example of four headlines for the same article that were created by
professional editors.
In this example, `電動車'(Electric cars) and `全'(all) are represented by red letters and are not included in the 24-character headline. These tokens cannot be evaluated by 24-character headlines.
The blue tokens are not included in 9- and 13-character headlines. These tokens should not be included in shorter headlines.
\label{tab:task}}
\end{table}

The news media publish newspapers in print form and in electronic form.
In the electric form, articles might be read on various types of devices using any application; thus, news media companies have an increasing need to produce multiple headlines for the same news article based on what would be most appropriate and most compelling on an array of devices. All devices and applications used for viewing articles have strict upper bounds regarding the number of characters allowed because of limitations in the space where the headline appears. 
The technology of automatic headline generation has the potential to contribute greatly to this domain, and the problems of news headline generation have motivated a wide range of studies~\citep{DBLP:conf/ijcai/WangYTZLD18, DBLP:conf/naacl/ChenLRLZLZ18, W18-5410, DBLP:conf/aaai/ZhouYWZ18, DBLP:conf/acl/LiWLC18, DBLP:conf/acl/WangQW19}. 

Table~\ref{tab:task} shows sample headlines in three different lengths written by professional editors of a media company for the same news article: The length of the first headline for the digital media is restricted to 10 characters, the second to 13 characters, and the third to 26 characters.
From a practical perspective, headlines must be generated under a rigid length constraint.

The first study to consider the length of system outputs in the encoder-decoder framework was \newcite{rush-chopra-weston:2015:EMNLP}. This study controlled the length of an output sequence by reducing the score of the end-of-sentence token to $-\infty$ until the method generated the desired number of words. 
Subsequently, \newcite{kikuchi-EtAl:2016:EMNLP2016} and \newcite{W18-2706} proposed mechanisms for length control; however, these studies produced summaries of 30, 50, and 75 bytes, and the studies evaluated the summaries by using the reference summaries of a single length (approximately 75 bytes long) in DUC 2004\footnote{\url{https://duc.nist.gov/duc2004/}}.
In addition, \newcite{takase-okazaki-2019-positional} proposed the mechanism for length control and evaluated their method with part of the test set which is consisted by summaries satisfying some length constraints in Annotated English Gigaword corpus~(AEG)~\citep{Napoles:2012:AG:2391200.2391218}. 

Thus, some questions can be posed: (1) Can previous evaluation settings adequately evaluate system outputs in headline generation task?
(2) What type of problem should we solve in this task according to the target length?
(3) How well do systems solve the problems?
In this study, we present novel corpora to investigate these research questions. The contributions of this study are threefold.
\begin{enumerate}
    \item We release the {\bf J}apanese {\bf N}ews {\bf C}orpus (JNC)\footnote{\url{https://cl.asahi.com/api_data/jnc-jamul-en.html}}, which includes 1.93 million pairs of headlines and the lead three sentences of Japanese news articles. We expect this corpus to provide common supervision data for headline generation.
    \item We build the {\bf JA}panese {\bf MU}lti-{\bf L}ength Headline Corpus (JAMUL)\footnotemark[2] for the evaluation of headlines of different lengths. In this novel dataset, each news article is associated with multiple headlines of three different lengths.
    \item We report new findings for the JAMUL; for example, although the longer reference seems to be able to evaluate the short system output, we also found a problem with this evaluation setting. Additionally, we clarified that the existing methods could not capture what must be changed according to the specified length.
\end{enumerate}

\section{JNC and JAMUL}

\begin{figure}[t]
	\centering
    \includegraphics[width=75mm]{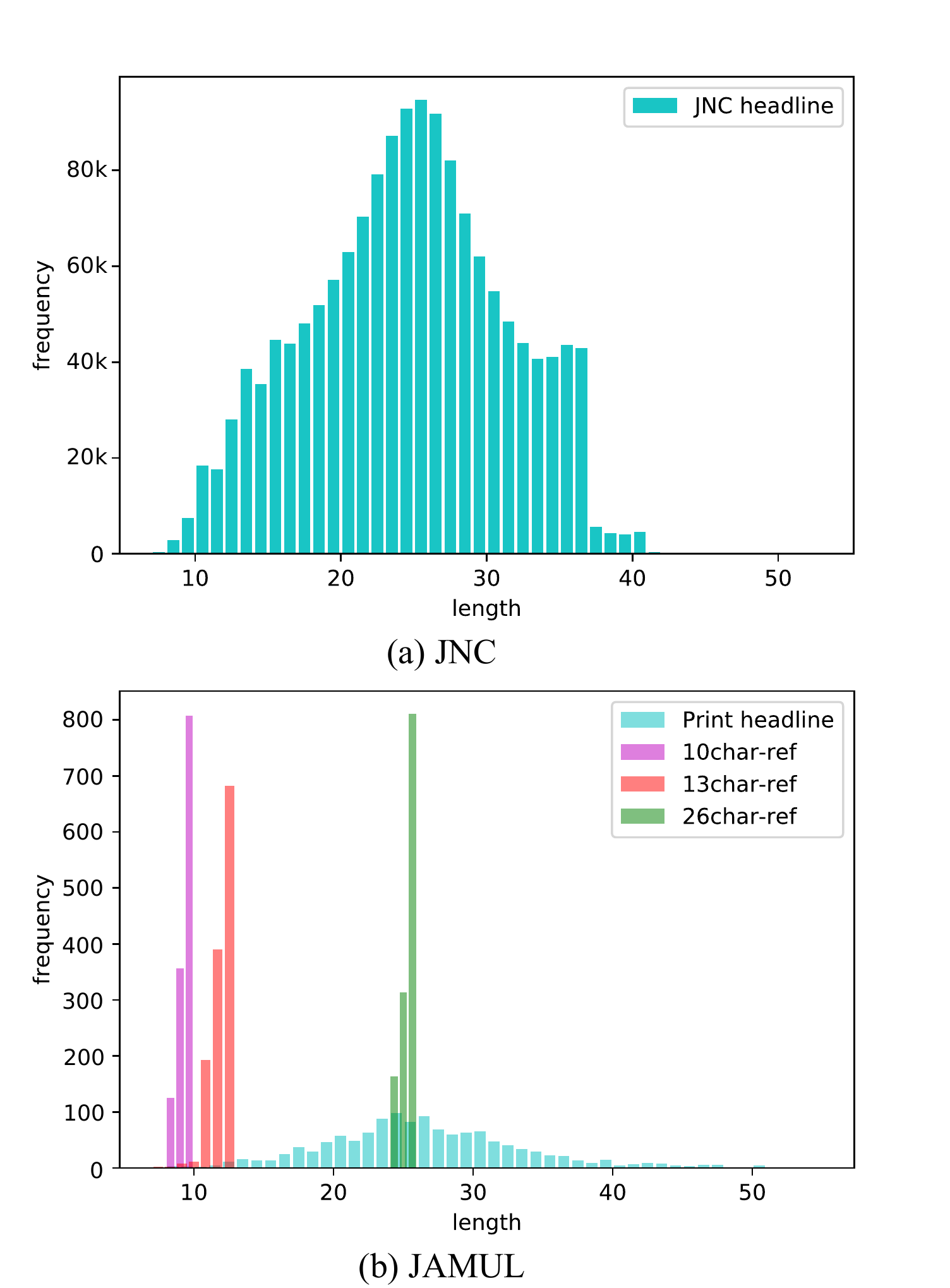}
	\caption{Length distributions of headlines in (a) the JNC and (b) the JAMUL.}
	\label{fig:testset}
\end{figure}

\subsection{Headlines Composed by a Media Company}

Before describing the JNC and the JAMUL in detail, we explain the process where a Japanese media company composes headlines for a news article. We describe the process of The Asahi Shimbun Company, a Japanese newspaper company.
First, reporters write an article and submit it to the editorial department. The editorial department picks up some articles to publish in the newspaper and writes a headline for the article dedicated to print media.
We call these headlines \emph{print headlines} or \emph{length-insensitive headlines} hereafter.

In addition to print headlines, digital media editors, who are typically not the same editors for print, pick up the articles they want to distribute on digital media from the articles dedicated to digital media and compose three different headlines. The first headline for the digital signage and audio media has a limit of up to 10 characters. 
This type of headline is appended to the beginning of a concise summary of the article so that readers can understand the news at first glance. The second type of headline is produced for mobile phones with small LCDs and small areas on the website sidebar (e.g., the access ranking); the upper limit of the number of characters is 13.
The third type of headline is produced for desktop computer news websites, and the upper limit of the number of characters is 26. This limit is derived from the layout of the news site.
We refer to the three types of headlines as \emph{10char-ref}, \emph{13char-ref}, and \emph{26char-ref}
 (refer to Table~\ref{tab:task} for examples).
We collectively call these headlines \emph{length-sensitive headlines}.

Table~\ref{tab:task} presents an example of headlines written for an article by the professional editors.
We extract the JNC and the JAMUL from the news production process of The Asahi Shimbun Company; therefore, they can be considered representative of contemporary editorial practice.

\subsection{JNC}
The JNC is a collection of 1,932,398 pairs of the three lead sentences\footnotemark[2] of articles and their print headlines published by The Asahi Shimbun Company from 2007 to 2016.
Figure~\ref{fig:testset} (a) depicts the distribution of the headline lengths in the JNC.
The headline lengths in the JNC are diverse because almost all headlines are not restricted by length limitations.

The JNC is useful for training headline generation models because it has many training instances. Furthermore, the corpus is suitable for training a model for variable-length headline generation because of the variety of the headline lengths.

\subsection{JAMUL}

The JAMUL is a corpus containing 1,489 full-text news articles and their length-insensitive headlines for print media and length-sensitive headlines of 10 characters, 13 characters, and 26 characters for digital media.
We just extract these articles and headlines from the company database.
All the articles and headlines were published by The Asahi Shimbun Company between September 2017 and March 2018.
The volume of the news articles may be insufficient for training a headline generation model.
However, as Figure \ref{fig:testset} (b) shows, the JAMUL includes length-sensitive headlines that strictly preserve the length requirements.
This novel characteristic of the JAMUL is suitable for the test set to adapt headline generation to business practice.
In this paper, we use the first three sentences of the  article as input because of fitting in the JNC corpus.
But we publish the full edition of the article. 
No overlap of articles between the JNC and the JAMUL was confirmed.
\subsection{Comparing of Headlines with Article Bodies}
\begin{table}[tb]
\centering
\small
\begin{tabular}{llrr}
\bhline{1pt}
System  & Reference                     & \multicolumn{1}{c}{Precision} & \multicolumn{1}{c}{Recall}  \\ \hline
\multirow{4}{*}{JAMUL Article} & Paper headline       & 11.26 & 72.91   \\
& 10char-ref           & 5.42  & 85.02  \\ 
& 13char-ref           & 7.10  & 85.74   \\ 
& 26char-ref           & 13.26 & 77.38   \\ \hline
AEG Article & AEG headline & 17.74 & 62.03 \\
\bhline{1pt}
\end{tabular}
\caption{
ROUGE-1 precision and recall scores when comparing article and length-insensitive/sensitive headlines.
\label{tab:article_coverage}}
\end{table}
What type of operation which includes extractive and abstractive operations did the editors perform to create length-sensitive and length-insensitive headlines in the JAMUL?
To clarify this question, we analyzed the proportions of the number of extractive and abstractive operations.
Specifically, we reported ROUGE-1 precision and recall scores~\citep{lin:2004:ACLsummarization} in Table~\ref{tab:article_coverage}, assuming that articles are ``system'' summaries and that 10char-ref, 13char-ref, and 26char-ref headlines are ``reference'' summaries.
Notably, we removed blank spaces, which were the most common token in longer headlines.
The relatively high recall score indicates that the operations most often required to generate headlines are extractive, and the abstractive operation is about 15--27\% of the total. In addition, we explored the proportions of each operation in AEG\footnote{We obtained this dataset by applying the pre-processing script at \url{https://github.com/facebookarchive/NAMAS}}, which has been used in many studies for English headline generation tasks. From the relatively high recall score, we observed that Japanese headlines tend to be more extractive.

\subsection{Comparing of Length-sensitive Headlines with Print headlines}
\begin{table}[tb]
\centering
\small
\begin{tabular}{llrr}
\bhline{1pt}
System  & Reference                     & \multicolumn{1}{c}{Precision} & \multicolumn{1}{c}{Recall}  \\ \hline
\multirow{3}{*}{Print headline} & 10char-ref     & 27.95    & 71.28     \\ 
 & 13char-ref     & 37.70    & 74.48    \\ 
 & 26char-ref     & 61.78    & 60.56    \\ \bhline{1pt}
\end{tabular}
\caption{
ROUGE-1 precision and recall scores when comparing length-insensitive and -sensitive headlines.}\label{tab:print_compare}
\end{table}

\begin{table}[t]
\centering
\small
\begin{tabular}{p{7.2cm}c}
\bhline{1pt}
{\bf Article: }米フェイスブック(FB)は1日、2017年7~9月期決算を発表し、モバイル広告の伸びなどで売上高、純利益ともに四半期として過去最高を記録した。\dots FB上で偽ニュースの拡散防止など、安全確保のための要員を約2万人に倍増させることを明らかにしている。\\
On the 1st U.S. Facebook (FB) announced financial results from July to September in 2017 and achieved a record quarterly amount of sales and net income thanks to its mobile advertising growth and other factors. \dots FB revealed it had doubled the number of personnel engaged in preventing fake news from spreading to about 20,000 to provide safety on FB.\\
\bhline{1pt}
{\bf Headline for print media: }フェイスブック、四半期で最高益　モバイル広告好調 \\
Facebook achieved a record quarterly profit thanks to its mobile advertising business. \\
\bhline{1pt}
{\bf Multi-length headlines for digital media: } \\
\begin{tabular}{cp{5.1cm}}
7 chars & 米FBが最高益 \\
(10char-ref)& U.S. FB achieved record profit. \\
12 chars & 米フェイスブックが最高益 \\
(13char-ref) & U.S. Facebook achieved record profit.\\
\multirow{5}{*}{ \shortstack{26 chars\\(26char-ref)} }& 米フェイスブックが最高益　偽ニュース対策で要員倍増も\\
		 & U.S. Facebook achieved record profit and doubled its personnel for countermeasures against fake news.\\
\end{tabular} 
\\
\bhline{1pt}
\end{tabular}
\caption{A typical example for comparing length-insensitive and -sensitive headlines
\label{tab:compare_print_digital}}
\end{table}

How similar are the headlines used for training (length-insensitive) and for evaluation (length-sensitive)?
We estimated the appropriateness of length-insensitive headlines as a ``seed'' for producing length-sensitive headlines.
More concretely, we report ROUGE-1 precision and recall scores in Table \ref{tab:print_compare}, assuming that length-insensitive headlines are ``system'' summaries, and that 10char-ref, 13char-ref and 26char-ref headlines are ``reference'' summaries.
The relatively high recall scores indicate that the training and evaluation data were not very distant.
Additionally, we found that the editors used a moderate number of words that did not appear in print headlines when composing length-sensitive headlines.
Table~\ref{tab:compare_print_digital} is an example of the typical differences between the length-insensitive and length-sensitive headlines.
Comparing the 26-character headline with the print headline, the choices of contents are different; for example, the print headline reports the reason for the record profit, but the 26-character headline describes the topic regarding the increasing number of personnel.
Next, comparing the 7-character (10char-ref) headline with the print headline, we observe that the choices of words are different; the print headline uses ``Facebook,'' which is changed to ``FB'' in the 7-character headline.

\subsection{Comparing of Length-sensitive Headlines}\label{seq:jamul_property}
\begin{table}[t]
\centering
\small
\begin{tabular}{llrr}
\bhline{1pt}
System  & Reference                     & \multicolumn{1}{c}{P} & \multicolumn{1}{c}{R}  \\ \hline
26char-ref & 10char-ref                       & 30.22     & 81.77  \\ 
26char-ref & 13char-ref                       & 43.59     & 90.75  \\ 
First 10 chars in 26char-ref & 10char-ref     & 44.30     & 29.42     \\ 
First 13 chars in 26char-ref & 13char-ref     & 67.01     & 44.34    \\
\bhline{1pt}
\end{tabular}
\caption{
Difference between 26char-ref headlines and shorter headlines. P and R denote ROUGE-1 precision and recall scores, respectively.
\label{tab:length-sensitive-compare}}
\end{table}

How similar is the composition of headlines for a news article of different lengths?
How good are 26char-ref headlines as ``seeds'' for generating 10char-ref or 13char-ref headlines?
Is the simple strategy of trimming 26char-ref headlines to 10 or 13 characters sufficient?
To answer these questions, we computed word-level precision and recall scores, assuming that 26char-ref headlines are ``system'' summaries and that 10char-ref and 13char-ref headlines are ``reference'' summaries.

The first and second rows of Table \ref{tab:length-sensitive-compare} represent the situation when we used 26char-ref headlines as they are, and without preserving the length constraint. 
Although this setting was unrealistic, we could estimate how similar the contents were among the length-sensitive headlines.
The high recall scores indicate that 26char-ref headlines mostly cover the words included in the 10char-ref and 13char-ref headlines.
From this result, we consider the contents among length-sensitive headlines are comparably similar.
The third and fourth rows of Table~\ref{tab:length-sensitive-compare} correspond to the strategy where we generated headlines in 10 and 13 characters from the first 10 and 13 characters of the 26char-ref headlines. 
This strategy achieved moderate success for generating headlines in 13 characters, but did not work well for headlines in 10 characters.
In other words, we observed large differences between the 10char-ref and the first 10 characters of the 26char-ref headlines. Therefore, we need to re-generate the shorter headline, as well as the longest one.

In sum, we found similarities in headlines of different lengths in the JAMUL. However, the simple strategy of trimming a longer headline into a shorter headline is insufficient.
Table~\ref{tab:task} is an example of the typical differences among length-sensitive headlines.
There is a little overlap between longer and shorter headlines because the 9- and 13-character headlines extract the shorter phrases which have the nearly same meaning as the 24-character headline.
Focusing on ``車種'' (models), the words are in the latter half of the 24-character headline, and we could confirm that important keywords are not always included at the beginning of the headlines.

\section{Comparing of Headline Generation Methods on the JAMUL}
In this section, we explore a question about evaluation: How reliable is the two conventional evaluation method; the first is the method that uses a single length summary for measuring the quality of summaries of different lengths; the second is the method that uses the specified length headlines which are extracted from the dataset constructed by length-insensitive headlines, for measuring the quality of each length summary.
To answer this question, we generated multiple summaries of different lengths by using the existing methods, and measured the correlation between the performance values computed by the conventional evaluation methods and those computed on the JAMUL.

\subsection{Headline-generation Methods with the Mechanism to Control the Output Length}\label{seq:models}
In this study, we explored four methods for headline generation that can control the output length. The first two methods, {\it LenEmb} and {\it LenInit}, were proposed by~\newcite{kikuchi-EtAl:2016:EMNLP2016}.

{\it LenEmb} provides the decoder with output length information in the form of the length embedding. {\it LenInit} controls the output length by multiplying the initial state of the decoder's memory cell of long short-term memory~\citep{DBLP:journals/neco/HochreiterS97} by the desired length.

\newcite{W18-2706} also proposed a length-controllable method for a convolutional sequence-to-sequence (ConvS2S) model~\citep{DBLP:conf/icml/GehringAGYD17}. Their method added special tokens indicating the range of the output length at the beginning of an input sequence.
In our experiment, we used a special token to specify an output length\footnote{\newcite{W18-2706} also included special tokens for entities, but we did not use them in the experiments.} and called this method {\it SP-token}.

We also considered the {\it LC} method~\citep{D18-1444}, which extends ConvS2S and multiplies the initial state of the residual connection~\citep{DBLP:conf/cvpr/HeZRS16} by the desired number of output tokens. In the experiment, we set the desired number of characters instead of that of tokens.

In addition to these four methods, we combined {\it SP-token} not only on ConvS2S but also on Seq2Seq and Transformer~\citep{DBLP:conf/nips/VaswaniSPUJGKP17}. Eventually, we examined six combinations in total: (1)~Seq2Seq $+$ {\it LenEmb}, (2)~Seq2Seq $+$ {\it LenInit}, (3)~Seq2Seq $+$ {\it SP-token},  (4)~ConvS2S $+$ {\it SP-token}, (5)~ConvS2S $+$ {\it LC}, and (6)~Transformer $+$ {\it SP-token}.

\begin{table}[]
\centering
\small
\begin{tabular}{lccc}
\bhline{1pt}
                     & \multicolumn{1}{l}{Seq2Seq} & \multicolumn{1}{l}{ConvS2S} & \multicolumn{1}{l}{Transformer} \\ \hline
Num of Layer           & 2                            & 8                         & 6                                \\ 
Dropout Rate         & 0.3                          & 0.1                       & 0.3                              \\
Grad Clipping     & {[}-5.0, 5.0{]}              & {[}-0.1, 0.1{]}           & -                                \\
Learning Rate       & 0.001                         & 0.2                       & 0.001                               \\
Optimizer           & Adam                         & NAG                       & Adam                                \\ \bhline{1pt}
\end{tabular}
\caption{
Parameters of each encoder-decoder model.
\label{tab:params}}
\end{table}

\begin{table*}[t]
\centering
\small
\begin{tabular}{llrrrrrrrrr} \bhline{1pt}
                         & \multicolumn{3}{c}{10 characters} & \multicolumn{3}{c}{13 characters} & \multicolumn{3}{c}{26 characters} \\ \hline 
Models                     & \multicolumn{1}{c}{R-1}      & \multicolumn{1}{c}{R-2}    & \multicolumn{1}{c}{R-L}     & \multicolumn{1}{c}{R-1}      & \multicolumn{1}{c}{R-2}    & \multicolumn{1}{c}{R-L}     & \multicolumn{1}{c}{R-1}     & \multicolumn{1}{c}{R-2}    & \multicolumn{1}{c}{R-L}     \\ \hline
(1) Seq2Seq + {\it LenEmb}      
                                 & 34.62    & 15.02  & 33.54   & 39.31   & 18.14  & 37.07   & 43.65   & 19.73  & 36.03   \\
(2) Seq2Seq + {\it LenInit}     & 36.14    & 16.88  & 35.15   & 41.39   & 19.55  & 38.83   & 46.30   & 21.33  & 37.79 \\
(3) Seq2Seq + {\it SP-token}    & 38.01    & 17.16  & 36.72   & 42.03   & 19.83  & 39.56   & 46.62   & 21.47  & 38.29   \\
(4) ConvS2S + {\it SP-token}    & 39.20    & 18.66  & 38.07   & 42.63   & 20.00  & 40.38   & 47.41   & 21.47  & 38.04   \\
(5) ConvS2S + {\it LC}          & 34.93    & 15.52  & 34.02   & 38.80   & 17.21  & 36.68   & 42.59   & 19.34  & 35.41   \\
(6) Transformer + {\it SP-token}& 41.21    & 19.16  & 39.91   & 44.84   & 21.62  & 42.09   & 49.66   & 23.69  & 40.65   \\ 
\bhline{1pt}
\end{tabular}
\caption{
ROUGE scores of each model on the JAMUL. The specified lengths are 10, 13, and 26 characters. R-1, R-2, and R-L represent ROUGE-1, ROUGE-2, and ROUGE-L, respectively. Note that (1) to (6) in Table~\ref{tab:duc_problem} and after that represent models (1) to (6) of this table.}
 \label{tab:JAMUL_eval}
\end{table*}

\begin{table*}[t]
\centering
\small
\begin{tabular}{lrrrrrr} \bhline{1pt}
& \multicolumn{3}{c}{10 characters} & \multicolumn{3}{c}{13 characters} \\ \hline 
Models & \multicolumn{1}{c}{R-1}      & \multicolumn{1}{c}{R-2}    & \multicolumn{1}{c}{R-L} & \multicolumn{1}{c}{R-1}      & \multicolumn{1}{c}{R-2}    & \multicolumn{1}{c}{R-L} \\ \hline
(1)          & $21.58_{[+0]}$ & $8.73_{[+1]}$  & $19.56_{[+0]}$ & $27.47_{[-1]}$ & $11.82_{[+0]}$  & $24.20_{[-1]}$\\
(2)          & $22.17_{[+0]}$ & $9.14_{[+1]}$  & $20.12_{[+0]}$ & $28.93_{[+1]}$ & $12.98_{[+2]}$  & $25.35_{[+1]}$ \\
(3)          & $22.80_{[+1]}$ & $9.51_{[+1]}$  & $20.59_{[+1]}$ & $28.78_{[-1]}$ & $12.64_{[-1]}$  & $25.14_{[-1]}$\\
(4)          & $22.53_{[-1]}$ & $9.04_{[-2]}$  & $20.47_{[-1]}$ & $29.23_{[+0]}$ & $12.82_{[-1]}$  & $25.58_{[+0]}$\\
(5)          & $21.60_{[+0]}$ & $8.55_{[-1]}$  & $19.63_{[+0]}$ & $27.63_{[+1]}$ & $11.72_{[+0]}$  & $24.34_{[+1]}$\\
(6)          & $24.36_{[+0]}$ & $10.32_{[+0]}$ & $21.97_{[+0]}$ & $30.84_{[+0]}$ & $13.85_{[+0]}$  & $26.98_{[+0]}$ \\ \hline
$\tau$  & 0.867  & 0.600 & 0.867  & 0.733 & 0.733  & 0.733 \\
\bhline{1pt}
\end{tabular}
\caption{ROUGE scores of the system outputs in 10 and 13 characters evaluated by 26char-ref headlines as the references. Note that [] denote the change of the rank when we compare the rank in Table~\ref{tab:JAMUL_eval}; + and - denote the rank up and rank down, respectively.}
 \label{tab:duc_problem}
\end{table*}

\begin{table*}[]
\centering
\small
\begin{tabular}{lrrrrrrrrr} \bhline{1pt}
& \multicolumn{3}{c}{10 characters} & \multicolumn{3}{c}{13 characters} & \multicolumn{3}{c}{26 characters} \\ \hline 
Models                     & \multicolumn{1}{c}{R-1}      & \multicolumn{1}{c}{R-2}    & \multicolumn{1}{c}{R-L}     & \multicolumn{1}{c}{R-1}      & \multicolumn{1}{c}{R-2}    & \multicolumn{1}{c}{R-L}     & \multicolumn{1}{c}{R-1}     & \multicolumn{1}{c}{R-2}    & \multicolumn{1}{c}{R-L}     \\ \hline
(1) & $37.95_{[+0]}$ & $19.97_{[+1]}$  & $36.54_{[+0]}$   & $41.54_{[+1]}$   & $21.60_{[+1]}$  & $39.41_{[+1]}$   & $33.97_{[+0]}$   & $13.54_{[-1]}$  & $27.66_{[+0]}$   \\
(2) & $38.35_{[-1]}$ & $20.71_{[+1]}$  & $37.04_{[-1]}$   & $41.52_{[-1]}$   & $21.13_{[-1]}$  & $39.05_{[-1]}$   & $35.50_{[+0]}$   & $14.18_{[+1]}$  & $28.64_{[+0]}$ \\
(3) & $41.35_{[+1]}$ & $21.95_{[+1]}$  & $39.93_{[+1]}$   & $43.62_{[+1]}$   & $22.51_{[+0]}$  & $40.89_{[+1]}$   & $35.66_{[+0]}$   & $14.14_{[-1]}$  & $28.80_{[+0]}$   \\
(4) & $39.33_{[-1]}$ & $20.47_{[-2]}$  & $37.46_{[-2]}$   & $40.38_{[-4]}$   & $20.47_{[-4]}$  & $37.53_{[-4]}$   & $36.84_{[+0]}$   & $14.42_{[+0]}$  & $29.21_{[+0]}$   \\
(5) & $39.21_{[+2]}$ & $19.59_{[-1]}$  & $37.69_{[+2]}$   & $42.99_{[+3]}$   & $22.94_{[+4]}$  & $40.53_{[+3]}$   & $33.36_{[+0]}$   & $13.55_{[+1]}$  & $27.59_{[+0]}$   \\
(6) & $45.14_{[+0]}$ & $26.35_{[+0]}$  & $43.70_{[+0]}$   & $47.09_{[+0]}$   & $25.80_{[+0]}$  & $43.95_{[+0]}$   & $37.86_{[+0]}$   & $15.68_{[+0]}$  & $30.46_{[+0]}$  \\ \hline
$\tau$  & 0.733 & 0.600 & 0.600 & 0.067  & -0.067 & 0.067  & 1.000 & 0.733 & 1.000 \\
\bhline{1pt}
\end{tabular}
\caption{
ROUGE scores of each model on the length-restricted JNC test set.}
 \label{tab:JNC_eval}
\end{table*}

\subsection{Datasets and the Evaluation Protocol}\label{japanesedataset}
We trained the six methods for headline generation on the JNC.
We removed instances that were duplicated or unsuitable for training a headline generation model\footnote{The filtering script is available at:\\ \url{https://github.com/asahi-research/Gingo}}.
The filtering step obtained 1,554,558 pairs of newspaper articles and headlines.
We randomly selected 98\% of the instances (1,523,468 pairs) as a training set, selected 1\% of the instances (15,545 pairs) as a validation set, and used the remainder (15,545 pairs) as a test set.
As for the JNC test set, we additionally extracted the length-restricted test set to evaluate the length control methods. Specifically, we extracted 303, 1,436, and 4,289 headlines which are 8-10, 11-13, and 24-26 characters long, respectively, and corresponding articles from the JNC, similar to the 10char-ref, 13char-ref, and 26char-ref in the JAMUL. We call 
them the {\it length-restricted JNC test set.}
We also filtered the JAMUL by setting lower bounds for the length of the headlines\footnotemark[6]. We set each lower bound to 8 for 10char-ref, to 11 for 13char-ref, and to 24 for 26char-ref. Finally, we achieved 1,288 instances for the JAMUL test set.

We used SentencePiece\footnote{\url{https://github.com/google/sentencepiece}}~\citep{kudo-richardson-2018-sentencepiece} for tokenization.
We set the merge operation to 8,000.
Finally, we obtained 10,868 tokens for the source side and 9,556 tokens for the target side.
When training a model, we set the length of each reference headline to the model. 
When generating headlines in the evaluation, we set the output lengths to 10, 13, and 26 characters; each output was evaluated by the reference that had the same length in the JAMUL and JNC test sets.
To generate summaries, we follow standard practice in disallowing repetition of the same trigram~\citep{DBLP:conf/iclr/PaulusXS18}.
We evaluated all models by using three variants of ROUGE recall metric\footnote{We used MeCab~\citep{kudo-yamamoto-matsumoto:2004:EMNLP} to tokenize the system outputs.}: ROUGE-1, ROUGE-2, and ROUGE-L.
Headlines exceeding the length limits were trimmed for a fair evaluation.

\subsection{Implementation}
We employed OpenNMT\footnote{\url{https://github.com/OpenNMT/OpenNMT-py}}~\citep{P17-4012} for Seq2Seq, and fairseq\footnote{\url{https://github.com/pytorch/fairseq}}~\citep{DBLP:conf/naacl/OttEBFGNGA19} for ConvS2S and Transformer.
We extended the implementations to realize {\it LenEmb, LenInit}, and {\it LC}.
We set the dimensions for the token and length embeddings to 512, those for hidden states to 512, and the beam width to 5.
These parameters are common in all the models. Table~\ref{tab:params} summarizes other parameters specific to each sequence-to-sequence model.
We used Nesterov's accelerated gradient method (NAG)~\citep{DBLP:conf/icml/SutskeverMDH13} with a momentum of 0.99 in ConvS2S.
In Transformer, we set the number of attention heads to 8, the dimensions for the feed-forward network to 2,048, Adam's $\beta$ to 0.98, the warm-up steps to 4,000, and label smoothing to 0.1.

\subsection{Evaluation of Multi-length Headlines Generated by Methods on the JAMUL}\label{seq:analysis1}
Table \ref{tab:JAMUL_eval} presents ROUGE scores of each method on the JAMUL test set.
Transformer + {\it SP-token} was the clear winner in all length and evaluation metrics.
Additionally, the three methods with {\it SP-token} outperformed the others.

What if we do not have multiple headlines of different lengths to evaluate the methods?
To answer this question, we followed the evaluation setup of the previous studies on DUC 2004: The reference summaries of 75 bytes were used even when evaluating summaries of 30 and 50 bytes. Table~\ref{tab:duc_problem} reports ROUGE scores for the system outputs in 10 and 13 characters evaluated based on the 26char-ref headlines.
This evaluation setup reduced the performance differences between the methods.
Although Transformer + {\it SP-token} remained the clear winner, the ranking in ROUGE scores of the other methods are flipped.
We also computed rank correlation coefficients (Kendall's $\tau$) to assess the discrepancy in the ranking among the methods presented in Table \ref{tab:JAMUL_eval} and Table~\ref{tab:duc_problem}.
The last row of Table \ref{tab:duc_problem} reveals that the rank correlation is not perfect (lower than one) but moderate. 
We understand that $\tau$ is maintained high to some extent because of two reasons: (1) Most of the swaps occur in adjacent ranks, and (2) there are not many samples of the method in this analysis. Then, we focus on the samples flipped in this evaluation setup, as shown in Table~\ref{tab:duc_problem}. We observed that the rankings are swapped even if there is a difference of 1.0 or more (these differences are well observed in state-of-the-art competition) in Table \ref{tab:JAMUL_eval}. For example, we can show this issue between (3) and (4) in 10 characters of R-1 score, (2) and (4) in 10 characters of R-2 score, and (3) and (4) in 10 characters setting of R-L score in Table~\ref{tab:JAMUL_eval}.

What if we do not have strict length headlines to evaluate the methods?
To answer this question, we followed the evaluation setup of previous studies : The reference summaries of each target length were extracted from the JNC test set.
Table~\ref{tab:JNC_eval} reports ROUGE scores for the system outputs in 10, 13, and 26 characters evaluated in the length-restricted JNC test set.
In this evaluation setup, Transformer + {\it SP-token} remained the clear winner, but the performances of the other methods were flipped.
We also calculated $\tau$ in this case.
The last row of Table~\ref{tab:JNC_eval} reveals that the 13 character setting considerably lost correlation with the JAMUL evaluation result. We guess this inconsistency is brought about by using the different content test sets among each length setting. 
Depending on the target length which is specified by the evaluation setting, this evaluation protocol might not evaluate the methods adequately.

\section{Analysis}

\subsection{Performance of Word Selection According to the Output Length}
How well do the existing methods change the word selection depending on the output length? As shown in the first and second rows of Table~\ref{tab:length-sensitive-compare}, the 10char-ref and 13char-ref headlines contain words that are not included in the 26char-ref headlines. In other words, the selection of words in the generated headline should be changed in response to the length restriction. To confirm this question, we computed ROUGE-1 recall scores for the system outputs generated by each method, assuming that the groups of the words included in the 10char-ref or 13char-ref but not in the 26char-ref headlines are the ``reference'' summaries. For instance, the red words in Table~\ref{tab:task} are the ``reference'' summaries in this experiment.

\begin{figure}[t]
	\centering
    \includegraphics[width=76mm]{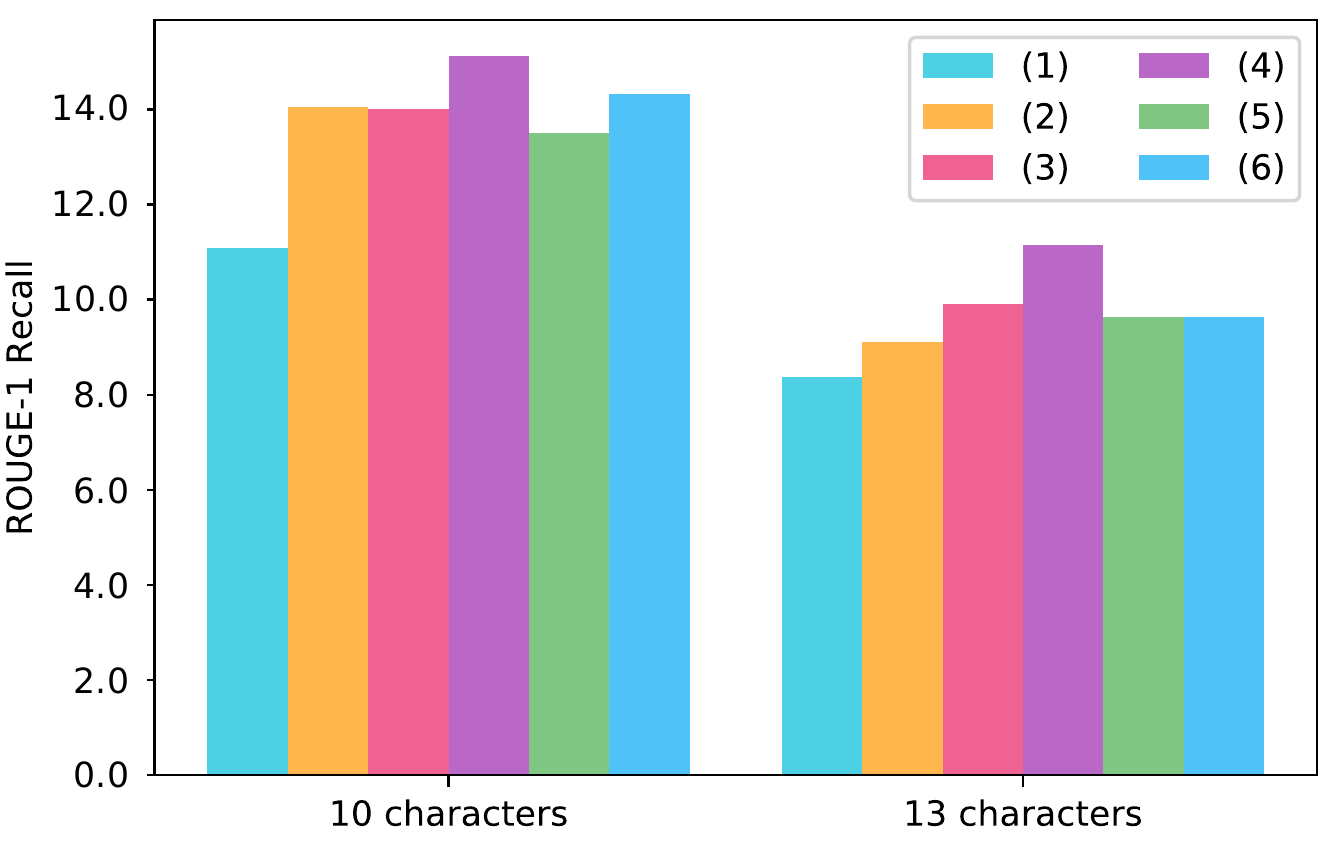}
	\caption{ROUGE-1 recall scores when comparing system outputs and the groups of the words included in 10char-ref or 13char-ref but not included in 26char-ref.}
	\label{fig:uniq_word}
\end{figure}

We report this result in Figure~\ref{fig:uniq_word}. The low recall score indicates that each system cannot select the words tailored to the length constraints.
However, (4) ConvS2S + {\it SP-token} shows the highest performance.
Taking this into account, Transformer + {\it SP-token} improves the important words in all lengths setting, but ConvS2S + {\it SP-token} may improve the change of words according to the target length.

\subsection{Performance of Managing Extractive and Abstractive Tasks}
In Table~\ref{tab:article_coverage}, we reported the proportion of the number of extractive and abstractive operations in the JAMUL. 
We analyze how the existing methods can reflect extractive and abstractive operations in generating summaries.

\begin{figure}[t]
	\centering
    \includegraphics[width=79mm]{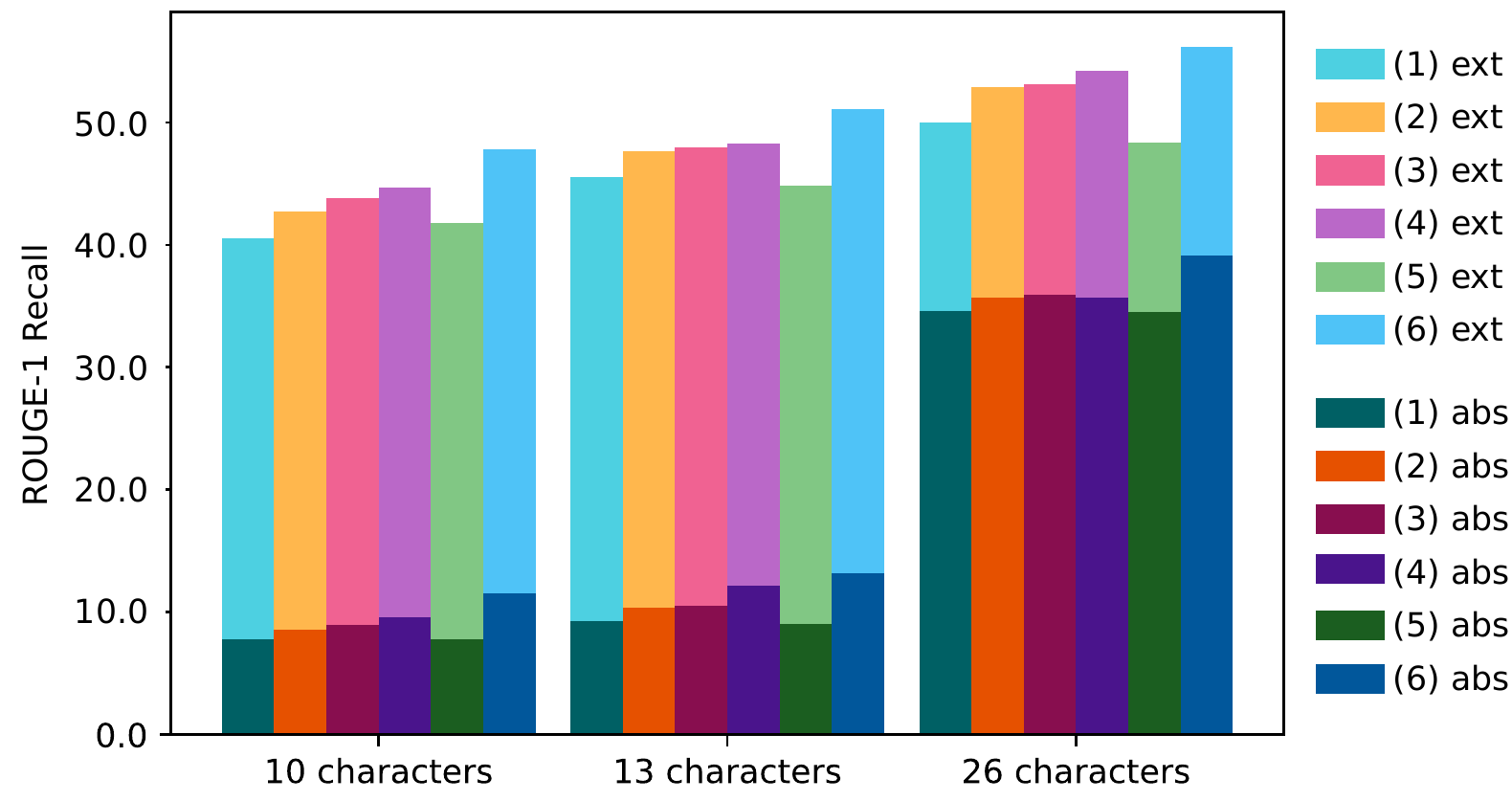}
	\caption{ROUGE-1 recall scores when comparing system outputs and the groups of the words which are extractive and abstractive.}
	\label{ext_abs_word}
\end{figure}

First, to observe extractive operations, we computed ROUGE-1 recall scores for the system outputs generated by each system, measuring the number of overlapping words between an article as ``system'' summaries and 10char-ref, 13char-ref, and 26char-ref headlines as ``reference'' summaries. The group of bars which contained {\it ext} in its name in Figure~\ref{ext_abs_word} reports the result.
The relatively high recall score indicates that the length control method succeeds in managing extractive operations.

Next, we examine whether the length control methods could perform abstractive operations. We adopted the words included in 10char-ref, 13char-ref, or 26char-ref headlines but not included in an article as ``reference'' summaries, and computed the ROUGE-1 recall scores for the system outputs. The group of bars which contained {\it abs} in its name in Figure~\ref{ext_abs_word} reports the result.
Regarding the outputs targeting 26 characters, the recall scores of around 30 point imply that each model can manage abstractive operations to some extent. In contrast, the low recall scores for the outputs targeting 10 and 13 characters revealed that all length control methods could not perform well on abstractive operations under the severe length constraint.

\subsection{How Do Length Control Mechanisms Work?}\label{seq:analysis2}
\begin{figure}[t]
	\centering
    \includegraphics[width=79mm]{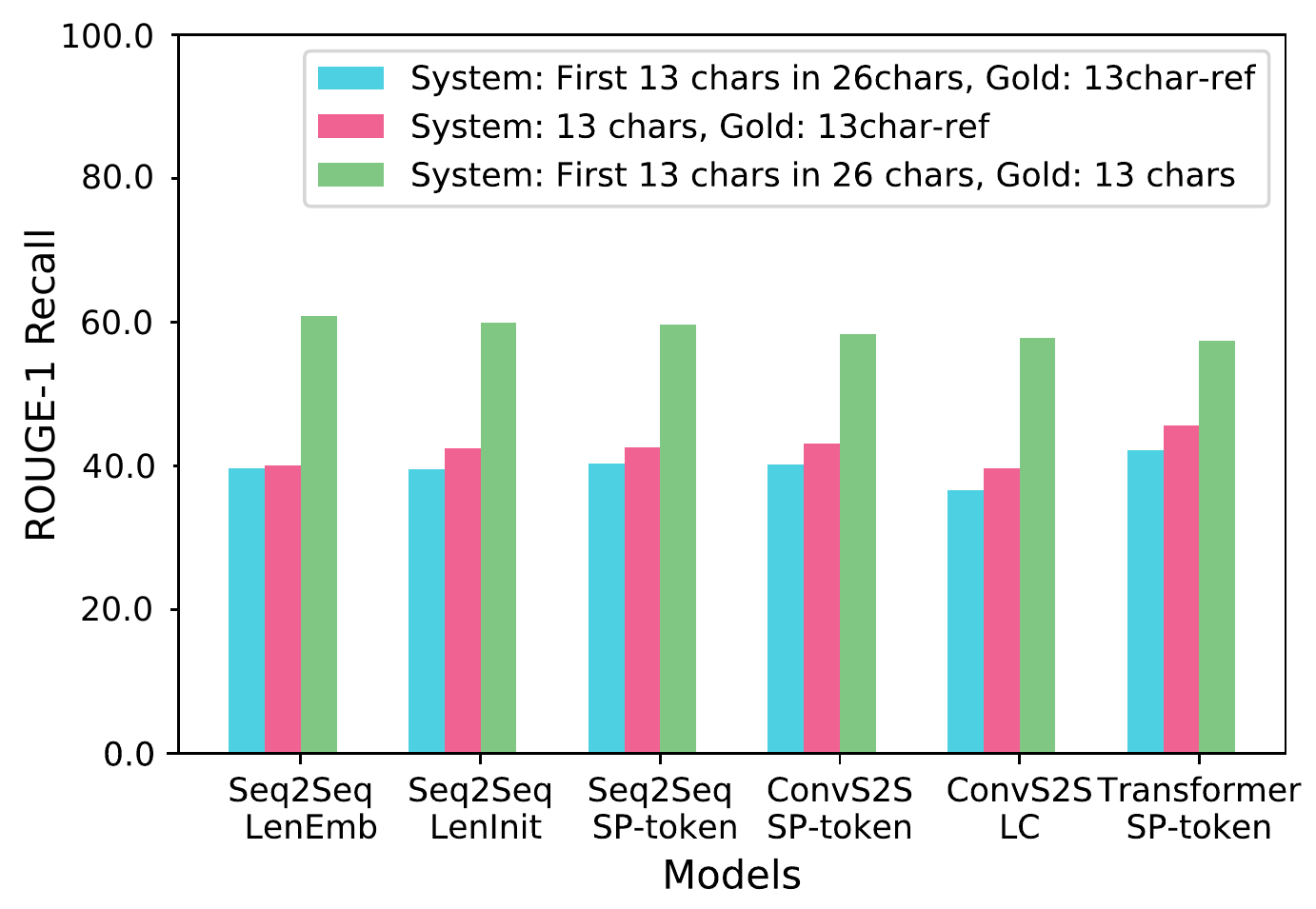}
	\caption{ROUGE-1 recall scores to assess the similarity of headlines generated for different lengths.}
	\label{seq:trim_same}
\end{figure}

We wondered whether a method that could control the output length would produce similar headlines even for different lengths for the same news article.
To confirm this suspicion, we reported ROUGE-1 recall scores in Figure~\ref{seq:trim_same} with three different configurations:
(a) evaluating the first 13 characters of headlines generated to be 26 characters long on 13char-ref headlines (blue);
(b) evaluating headlines generated to be 13 characters long on 13char-ref headlines (pink);
and setting (c) is the same as setting (a) but evaluated on the headline generated to be 13 characters long (green).

Setting (a) corresponds to the strategy where we trimmed headlines of different lengths to 13 characters long.
This setting was worse than setting (b), where a method tailored headlines to the desired length.
However, the difference in ROUGE scores between (a) and (b) was not very large, indicating that the existing methods do not drastically change the content for 13 characters long and 26 characters long.
This tendency was also verified by setting (c), which assessed how much the first 13 characters of headlines generated to be 26 characters long covered the content of those generated to be 13 characters long.
These facts suggest that we should explore in further research a method not only trained by generic supervision data (print headlines) but also tuned for the desired length.

\section{Related Work}

\newcite{rush-chopra-weston:2015:EMNLP} created the first approach to neural abstractive summarization.
They generated a headline from the first sentence of a news article in the AEG~\citep{Napoles:2012:AG:2391200.2391218}, which contains an enormous number of pairs of headlines and articles.
After their study, a number of researchers addressed this task: For example,~\newcite{chopra-auli-rush:2016:N16-1} used the encoder-decoder framework~\citep{Sutskever:2014:SSL:2969033.2969173,DBLP:journals/corr/BahdanauCB14} and~\newcite{DBLP:conf/conll/NallapatiZSGX16} incorporated additional features into the model, such as parts-of-speech tags and named entities.
\newcite{suzuki-nagata:2017:EACLshort} proposed word-frequency estimation to reduce the repeated phrases being generated. 
\newcite{P17-1101} proposed a gating mechanism ({\it sGate}) to ensure that important information is selected at each decoding step. \par
Furthermore, attempts to control the output length in neural abstractive summarization have been gradually increasing.
\newcite{shi-knight-yuret:2016:EMNLP2016} reported that hidden states in recurrent neural networks in the encoder-decoder framework could implicitly model the length of the output sequences. \newcite{kikuchi-EtAl:2016:EMNLP2016} was the first to propose the idea of controlling the output length in the encoder-decoder framework. Their approach inserts length information for the output length into the decoder. Additionally, \newcite{W18-2706} reported that output lengths could be controlled by embedding special tokens given to an input sequence. These two studies used DUC 2004~\citep{Over:2007:DC:1284916.1285157}, which comprises only 75-byte summaries, to evaluate the outputs in multiple lengths. \newcite{D18-1444} also proposed a method for controlling the number of output tokens in the ConvS2S model.
In Transformer~\citep{DBLP:conf/nips/VaswaniSPUJGKP17}, \newcite{takase-okazaki-2019-positional} proposed two length control methods by extending positional embedding. They additionally evaluated the system outputs by using the reconstructed test set of AEG which consists of the fixed length headlines.
\newcite{DBLP:conf/acl/MakinoITO19} proposed a global optimization method under a length constraint. \par
\newcite{sun-etal-2019-compare} examined how to compare summarizers by considering the length bias of generated summaries in the test set that includes various length summaries.
However, no previous work built a dataset for evaluating headlines of multiple lengths or reported an in-depth perspective on this task during the process of new production in the real world.
However, a single length reference that could appropriately evaluate multiple length summaries in multiple document summarization was reported~\cite{DBLP:conf/emnlp/ShapiraGRBAND18}.
In that study, the authors confirmed the correlation coefficient of ROUGE scores between the scores using a single length reference and multiple (gold) length references in the evaluation.
The present research differed in that we examined what kind of problems occurred and studied headline generation domain, which requires stricter keyword selection.

\section{Conclusion}
In this paper, we presented two new corpora: The JNC contains a large number of pairs of news articles and their headlines, and the JAMUL includes headlines of three different lengths (10, 13, and 26 characters long) written by professional editors. This study is the first to analyze the characteristics of multiple headlines of different lengths, and to evaluate existing approaches for length control based on the reference headlines composed for different lengths.
We found that Transformer model with a special length token ({\it SP-token}) outperformed the other methods on the JAMUL.
Additionally, although we confirmed that single length (the longest) references could adequately evaluate multiple length system outputs, the rankings were swapped even if there is a difference of 1.0 or more in the ROUGE scores.
We also confirmed the length-restricted test set, which was extracted from length-insensitive headlines and corresponding articles, could not adequately evaluate the multiple length system outputs depending on the specified length.
In the analysis, the existing methods could not take into account the word selection according to length constraint.
We also found it difficult to evaluate methods to controlling output length, because headlines of different lengths are written based on different goals, and because the training data does not necessarily reflect the goal of the headlines of a specific length.
In the future, we plan to explore an approach to adapt a model trained on print headlines to those which dedicated to a different length.

\bibliography{emnlp-ijcnlp-2019}
\bibliographystyle{acl_natbib}

\end{document}